\documentclass[11pt,a4paper]{article}
\usepackage{fullpage}
  \usepackage[pdftex]{graphicx}
\usepackage[cmex10]{amsmath}
\usepackage[sort,nocompress]{cite}
\usepackage{amssymb}
\usepackage[caption=false,font=footnotesize]{subfig}
\usepackage{stfloats}
\usepackage{hyperref}
\usepackage{framed}
\usepackage{color}
\usepackage[dvipsnames]{xcolor}
\usepackage{xspace}

\usepackage{multirow}

\newcommand{\sv}{\textsf{Structure2Vec}\xspace}
\newcommand{\sTv}{\textsf{MFE}\xspace}
\newcommand{\aTv}{\textsf{i2v\_mean}\xspace}
\newcommand{\wTv}{\textsf{i2v\_attention}\xspace}
\newcommand{\rnnTv}{\textsf{i2v\_RNN}\xspace}

\newcommand{\sampTv}{\textsf{Sampling}\xspace}
\newcommand{\normalDataset}{\textsf{OpenSSL Dataset}\xspace}

\newcommand{\simDataset}{\textsf{Qualitative Dataset}\xspace}

\newif\ifshowcomments
\showcommentstrue

\ifshowcomments
\newcommand{\mynote}[2]{\fbox{\bfseries\sffamily\scriptsize{#1}}
 {\small$\blacktriangleright$\textsf{\emph{#2}}$\blacktriangleleft$}}
\else
\newcommand{\mynote}[2]{}
\fi

\begin{document}
%
\title{Unsupervised Features Extraction for Binary Similarity Using Graph Embedding Neural Networks}
\author{Roberto Baldoni$^*$, Giuseppe Antonio Di Luna, Luca Massarelli$^*$, \\Fabio Petroni, Leonardo Querzoni$^*$.\\
	*:University of Rome Sapienza;\\
         \footnotesize{\{baldoni,massarelli,querzoni\}@diag.uniroma1.it, {g.a.diluna@gmail.com}, petronif@acm.org}}
\date{}

\maketitle

\begin{abstract}


In this paper we consider the binary similarity problem that consists in determining if two binary functions are similar only considering their compiled form. This problem is know to be crucial in several application scenarios, such as copyright disputes, malware analysis, vulnerability detection, etc.
Current state-of-the-art solutions in this field \cite{CCS} work by creating an embedding model that maps binary functions into vectors in $\mathbb{R}^{n}$. 
Such embedding model captures syntactic and semantic similarity between binaries, i.e., similar binary functions are mapped to points that are close in the vector space.
This strategy has many advantages, one of them is the possibility to precompute embeddings of several binary functions, and then compare them with simple geometric operations (e.g., dot product).


 
In \cite{CCS} functions are first transformed in Annotated Control Flow Graphs (ACFGs) constituted by manually engineered features and then graphs are embedded into vectors using a deep neural network architecture.
In this paper we propose and test several ways to compute annotated control flow graphs that use unsupervised approaches for feature learning, without incurring a human bias.
Our methods are inspired after techniques used in the natural language processing community (e.g., we use word2vec to encode assembly instructions). We show that our approach is indeed successful, and it leads to better performance than previous state-of-the-art solutions. Furthermore, we report on a qualitative analysis of functions embeddings. We found interesting cases in which embeddings are clustered according to the semantic of the original binary function.

\end{abstract}


%
\section{Introduction}
In the last years there has been an exponential increase in the creation of new contents. As all products, also software is subject to this trend. As an example, the number of apps available on the Google Play Store increased from 30K in 2010 to 3 millions in 2018\footnote{ \url{https://www.statista.com/statistics/266210/number-of-available-applications-in-the-google-play-store/}}.
This increase directly leads to more vulnerabilities as reported by CVE\footnote{\url{https://www.cvedetails.com/browse-by-date.php}} that witnessed a 120\% growth in the number of discovered vulnerabilities from 2016 to 2017. At the same time complex software spread in several new devices, the internet of things has multiplied the number of architectures on which the same program has to run and COTS software components are increasingly integrated in closed-source products.

This multidimensional increase in quantity, complexity and diffusion of software makes the resulting infrastructures difficult to manage and control, as part of their internals are often inaccessible for inspection to their own administrators.
As a consequence, system integrators are looking forward to novel solutions that take into account such issues and provide functionalities to automatically analyze software artifacts in their compiled form (binary code). One prototypical problem in this regard, is the one of {\em binary similarity} \cite{bindiff,khoo,sigma}, where the goal is to find pieces of software that are similar to each other according to some definition of similarity.

Binary similarity has been recently subject to a lot of attention \cite{David,David3,blanket}. This is due to its centrality in several tasks, such as discovery of known vulnerabilities in large collection of softwares, dispute on copyrights matters, analysis and detection of malicious software, etc.

In this paper, in accordance with \cite{genius} and \cite{CCS}, we focus on a specific version of the binary similarity problem in which we define two binary functions to be similar if they are derived from the same source code. As already pointed out in \cite{CCS}, this assumption does not make the problem trivial.
The complexity lies in the fact that, starting from the same source code, widely different binaries can be generated by different compilers with several optimisation parameters. To make things more challenging the same source code could be compiled targeting different architectures that use completely different instruction sets (in particular we consider AMD64 and ARM as target architectures for our study).

There are at least two important use cases for which this version of binary similarity is appropriate. The first one is the detection of known vulnerabilities: a vulnerable source code is usually well known, and it has to be searched against a, possibly really large, set of closed-source executables compiled by different organisations for different architectures. The second one is the phylogenic analysis of malware: in such scenario the source code is not known, but analysts may suspect a certain relationship between two malwares, that can be confirmed by finding shared source code.

Inspired by \cite{genius} we look for solutions that solve the binary similarity problem using {\em embeddings}. Loosely speaking, each binary function is first transformed in a {\em control flow graph} (CFG), where blocks of instruction representing vertices are connected by edges defining the execution flow among these blocks. Salients features are extracted from the CFG obtaining a new graph that is is mapped into a vector of numbers (an \emph{embedding}), in such a way that similar graphs result in vectors that are similar.

This idea has several advantages. First of all, once the embeddings are computed checking the similarity is relatively cheap and fast (we consider the scalar product of two constant size vectors as a constant time operation), thus we can pre-compute a large collection of embeddings for interesting functions and check against such collection in linear time. In the light of aforementioned use cases, this characteristic is extremely useful.
Another advantage comes from the fact that such embeddings can be used as input to other machine learning algorithms, that can in turn cluster functions, classify them, etc.


We compute binary function embeddings using \emph{Graph Embedding Networks}. These family of deep neural architectures, initially proposed in \cite{CCS}, consist of two main components: the first one extracts from each binary function an {\em Annotated Control Flow Graph} (ACFG) \cite{genius}; the second one takes in input this ACFG and use a deep neural network based on \sv \cite{dai2016discriminative} to generate an embedding vector for the graph. Each ACFG vertex contains features that are generally manually selected (e.g., the number of strings or the number of constants inside a certain block of source code). In this paper we argue that learning such vertex features in an unsupervised fashion can lead to advantage in performance with respect to manually engineering them.

In fact, a really succesfull recent trend in machine learning is to design network architectures that introduce the smallest possible human bias and that automatically learn the representations needed from raw data \cite{bengio2013representation,Mikolov}. We argue that manually selecting vertex features is prone to injecting a bias that is not supported by a thorough and formal a-priori investigation of the binary similarity problem (at the best of our knowledge there is no formal model justifying the centrality of the features selected in \cite{genius,CCS}). 

In this paper we investigate the suitability of an end-to-end trainable system that does not use humanly selected features in the context of the binary similarity problem.
In particular we argue that an unsupervised approach for feature learning would allow for an effective representation of each CFG vertex, without incurring a human bias. 
Such unsupervised feature learning approach takes inspiration from the natural language processing community in that it considers binary instructions as word tokens.
In particular, we learn embedding models that represents binary instruction as real-valued vectors and capture syntactic and semantic similarity between them.
We then leverage such instruction embedding models to generate a feature vector for each vertex in the CFG.

Summing up, the main contributions of our work are the following:
\begin{itemize}
	\item we describe a general network architecture for calculating binary function embeddings starting from the corresponding CFGs that extends the one introduced in \cite{CCS}; 
	\item we introduce several designs for the \emph{Vertex Features Extractor}, a component that associates a feature vector to each vertex in the CFG. Most of these designs make use of unsupervised learning techniques, i.e. they do not introduce any human bias;
	\item we report on an experimental evaluation conducted on these different feature extraction solutions considering the same dataset introduced in \cite{CCS};
	\item we provide some insights on the embedding vector space through a qualitative analysis.
\end{itemize}


The experiments confirmed that our claim was indeed true: removing human bias from the network architecture results in embeddings that provide better performance in the binary similarity context.

The remainder of this paper is organized as follows. Section \ref{sec:related_work} discusses related work, followed by Section \ref{sec:problem_definition} where we define the problem and report an overview of the solutions we tested. In Section \ref{sec:theodetails} we describe in details each solution. Finally in Section \ref{sec:evaluation} we describe the experiments we performed and report their results, as well as a qualitative analysis of the function embedding produced by our best performing model.

\section{Related Work}
\label{sec:related_work}

We can divide the binary similarity literature in works that propose and test solutions for a single architecture (e.g., AMD64), and works that design solutions for cross-architecture binary similarity. 

Regarding the literature of binary-similarity for a single architecture a family of works is based on on matching algorithms for the CFGs of functions. In Bindiff \cite{bindiff} the node matching is based on the syntax of code, and it performs poorly across different compiler, see \cite{David}. 
 In \cite{pewny} each vertex of a CFG is represented with an expression tree. Similarity among vertices is computed by using the edit distance between the corresponding expression trees.  
Other works are not based on graph matching: In  \cite{David2} the idea is to represent a function as several independent execution traces, called tracelets. Similar tracelets are then matched by using a custom edit-distance.  A related concept is used in \cite{David} where functions are divided in pieces of independent code, called strands. The matching between functions is based on how many statistically significant strands are similar. Intuitively, a strand is significant if it is not statistically common. The strand similarity is done using an smt-solver to assess the semantic similarity.  
Note that all previous solutions are designed around matching procedures that work \emph{pair-to-pair}, and they cannot be adapted to pre-compute a constant size signature of a binary function on which similarity can be assessed. 
Egele et al. in \cite{blanket} proposed a solution where each function is executed multiple times in a random environment. During the executions some features are collected and then used to match similar functions.
Note that this solution can be used to compute a signature for each function. However, it needs to execute a function multiple time, that is both time consuming and difficult to perform in the cross-architecture scenario.
Moreover, it is not clear if the features identified in  \cite{blanket} are useful for cross-architecture comparison.   
Finally, in \cite{khoo} by Khoo, Mycroft, and Anderson the matching is based on \emph{n-grams} computed on instruction mnemonics and graphlets. Even if this strategy does produce a signature, it cannot be immediately extended it to cross-platform similarity. 
 
\begin{table}[t!]
\caption{Notation.}	
\centering
\begin{tabular}{  r | l  }	
  $s$ & source code \\
  $c$ & compiler \\
  $f^{s}_{c}$ & binary function compiled from source code $s$ with compiler $c$ \\
  $g$ & Control Flow Graph (CFG) \\
  $\vec{g}$ & embedding vector of $g$ \\
  $V$ & set of vertices in $g$ \\
  $E$ & set of edges in $g$ \\
  $v_i$ & $i$-th vertex of $V$ \\
  $\mathcal{N}(v_i)$ & set of neighbours of vertex $v_i$ \\
  $x_i$ & features vector of vertex $v_i$ \\
  $d$ & dimension of feature vector $x_i$ \\
  $I_{v_i}$ &  list of instructions in the vertex $v_i$ \\
  $m$ & number of instructions in vertex $v_i$, that is $|I_{v_i}|$ \\
  $\iota$  & instruction in $I_{v_i}$ \\ 
  $\vec{\iota}$  & embedding vector of $\iota$ \\ 
  \textsf{i2v} & instruction embedding model (instruction2vector) \\
  $a$ & weights vector for \wTv \\
  $a[i]$ & $i$-th component of vector $a$ \\
  $h^{(i)}$ & $i$-th hidden state of a Recurrent Neural Network (RNN) \\
  $\mu_i$ & vector used within the Structure2vec algorithm associated with vertex $v_i$ \\
  $p$ & dimension of vector $\mu_i$  \\
  $\mu_i^t$ & state of vector $\mu_i$ at round $t$ \\
  $\mathcal{F}$ & nonlinear function, see Equation \ref{eq:f} \\
  $\sigma$ & nonlinear function, see Equation \ref{eq:sigma} \\
  $tanh$ & hyperbolic tangent function \\
  $\text{ReLU}$ & rectified linear unit function \\
  $W_1$ & $d \times p$ matrix - hyperparameter of Structure2vec \\
  $W_2$ & $p \times p$ matrix - hyperparameter of Structure2vec \\
  $P_i$ & $i$-th matrix of dimension $p \times p$ - hyperparameter of Structure2vec \\
  $\Pi$ & set of all possible vectors obtained by permuting the argument components\\
  $\ell$ & number of layers in Structure2vec and of $P$ matrices \\
  $T$ & total number of rounds in Structure2vec \\
  $K$ & number of input training pairs of CFGs \\
  $y_i$ & ground truth label associated with the $i$-th input pair of CFGs \\
  $\Phi$ & network hyperparameters \\
  $J$ & network objective function \\
\end{tabular}
\label{table:nota}
\end{table}

Regarding the literature of  cross-architecture binary-similarity.  In \cite{crossarch} a graph-based methodology is proposed, i.e. it used a matching algorithm on the CFGs of functions. The idea is to transform the binary code in an intermediate representation; on such representation the semantic of each CFG vertex is computed by using a sampling of the code executions using random inputs. 
Feng et al.~\cite{asiaccs} proposed a solution where each function is expressed as a set of conditional formulas. Then they used integer programming to compute the maximum matching between formulas.
Note that both, \cite{crossarch} and  \cite{asiaccs} are \emph{pair-to-pair}. 
David, Partush, and Yahav in \cite{David3} transformed binary code to an intermediate representation. Then, functions were partitioned in slices of independents code, called \emph{strands}. An involved process guarantees that strand with the same semantics will have similar representations. Functions are deemed to be similar if they have matching of significant strands. Note that this solution does generate a signature as a collection of hashed strands. However, it has two drawbacks, the first is that the signature is not constant-size but it depends on the number of strands contained in the function. The second drawback is that is not immediate to transform such signatures into vectors of real numbers that can be directly fed to other machine learning algorithms.

The most related to our works are the one that propose embeddings for cross-platform binary similarity.  
In 2016 Feng et al.~\cite{genius} introduced a solution that uses a clustering algorithm over a set of functions to obtains centroids for each cluster. Then, they used these centroids and a configurable feature encoding mechanism to associate a numerical vector representation with each function. According to the authors this approach outperforms several state of the art solutions in the cross-platform bug search. 
Embeddings were also used by Xu et al. in \cite{CCS} where they are computed using a deep neural network. Interestingly, \cite{CCS} shows that this strategy outperforms \cite{genius} both in terms of accuracy, performance (measured as time required to train the model), and flexibility (neural networks can be retrained to solve specific subtasks). To the best of our knowledge, these are the only two scientific works proposing embedding-based solutions for binary-similarity.


\section{Problem Definition and Solutions Overview}
\label{sec:problem_definition}
For clarity of exposition we summarize all the notation used in this paper in Table \ref{table:nota}.
We say that two binary functions $f^{s}_1,f^{s}_2$ are similar, $f_1 \sim f_2$, if they are the result of compiling the same original source code $s$ with different compilers. 
Essentially, a compiler $c$ is a deterministic transformation that maps a source code $s$ to a corresponding binary function $f^{s}_{c}$. In this paper we consider as a compiler the specific
software, e.g. gcc-5.4.0, together with the parameters that influence the compiling process, e.g.
the optimisation parameters -O$[0,...,3]$.

\begin{figure}

  \centering
    \includegraphics[width=\columnwidth]{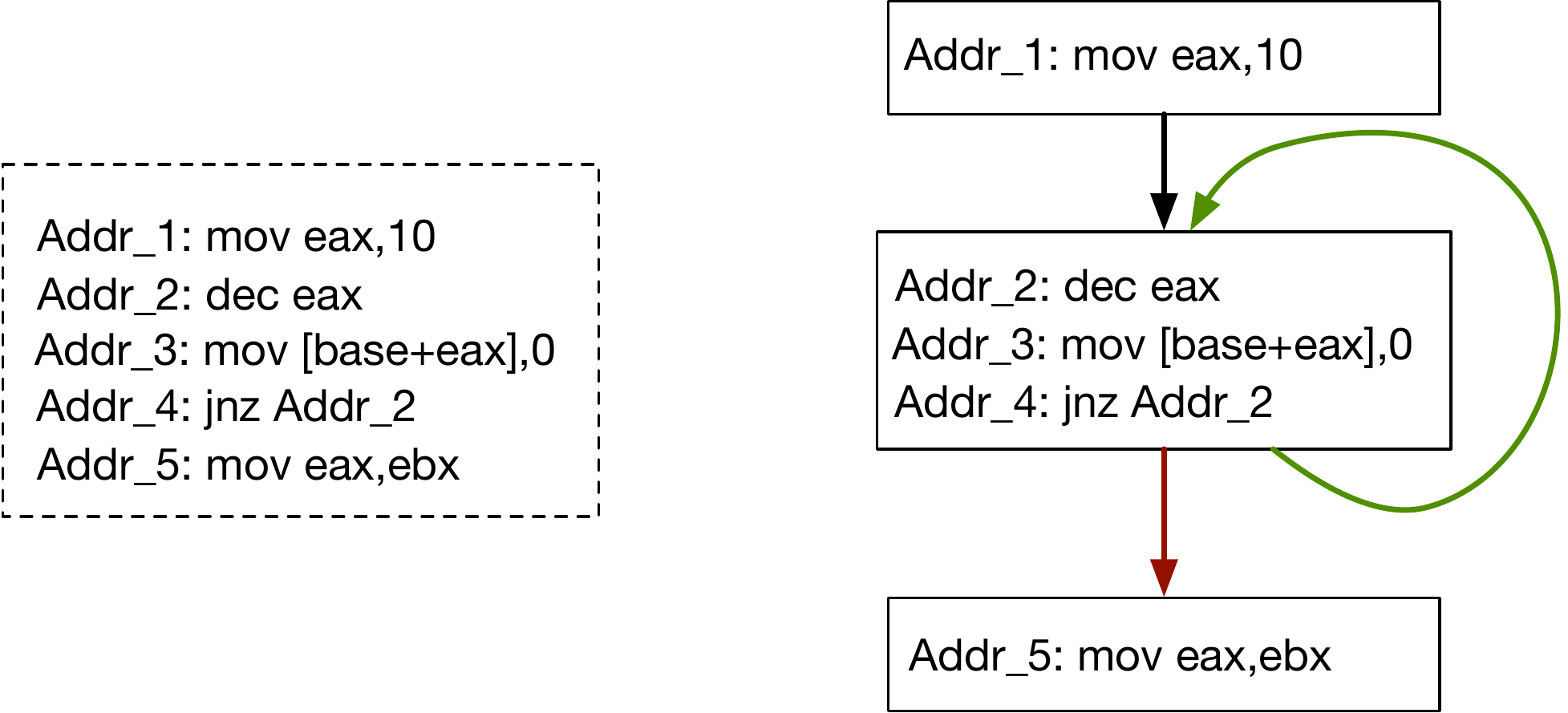}
      \caption{Assembly code on the left and corresponding cfg graph on the right. \label{fig:cfg}}
\end{figure}

 Given a binary function, we use its representation as {\em control flow graph} (CFG) \cite{allen}. A control flow graph is a directed graph whose vertices are sequences of assembly instructions, and whose arcs represent the execution flow among vertices. In Figure \ref{fig:cfg} there is a binary code snippet and the corresponding CFG. 
We reduce the binary similarity problem to the one of finding similar CFGs. Thus when we say that two CGFs are similar we mean that the corresponding functions are similar (the same hold for dissimilar CFGs). 
We denote a CFG as $g = ({V,E})$ where $V$ is a set of vertices and $E$ a set of edges. 
We denote as $\mathcal{N}(v_i)$ the set of neighbours of vertex $v_i \in V$. 
Vector $x_i$ is a $d$-dimensional features vector associated with vertex $v_i$, and $I_{v_i}$ is the set of instructions contained in vertex $v_i$.
Without loss of generality, we assume that all vertices contains the same number of instructions $m$\footnote{Concretely speaking this is achievable by adding NOP padding instructions to vertices that contains less than $m$ instructions.}.
Our aim is to represent a CFG as a vector in $\mathbb{R}^{n}$. 
This is achieved with an embedding model that maps a CFG $g$ to an {\em embedding vector} $\vec{g} \in \mathbb{R}^{n}$, preserving structural similarity relations between CFGs, therefore between binary functions.
We use a graph embedding neural network as embedding model.
%
The first component of this network is a {\em Vertex Features Extractor} mechanism that automatically generates a feature vector $x_i$ for a vertex $v_i$ in the CFG using the set of instructions $I_{v_i}$.

In this paper we consider three general approaches to implement the vertex feature extractor component: (A) manual feature engineering, (B) unsupervised feature learning and (C) instructions execution and result sampling.

\subsection{Manual Feature Engineering (\sTv)}
This is the approach defined in \cite{CCS}. The feature vector $x_i$ of vertex  $v_i$ is computed by counting the number of instructions belonging to predefined classes (as example, transfer instructions), together with number of strings and constants referred in $v_i$, and the offspring and betweenness centrality of $v_i$ in the CFG. Note that this mechanism is not modified by the training procedure of the neural network, that is is not {\em trainable}. This approach is our baseline.

\subsection{Unsupervised Feature Learning} 
The main idea of this family of approaches is to map each instruction $\iota \in I_{v_i}$ to vectors of real numbers $\vec{\iota}$, using the word2vec model \cite{Mikolov}, an extremely popular feature learning technique in natural language processing. We use a large corpus of instructions to train our instruction embedding model (see Section \ref{sec:implementation}) and we call our mapping instruction2vec (\textsf{i2v}).
Note that such instruction vectors can be both kept {\em static} throughout training or updated via backpropagation ({\em non-static}) by the network.
To generate a single feature vector $x_i$ for each vertex $v_i$ we considered three different strategies to aggregate the instructions embedding:
\begin{itemize}
\item \aTv: the feature vector $x_i$ of vertex $v_i$ is obtained by computing the arithmetic mean of the vectors $\vec{\iota}$ of instructions $\iota \in I_{v_i}$.
{\bf Rationale:} this is a simple but effective and popular way to aggregate embedding vectors \cite{arora2016simple,nugent2017comparison}.
\item \wTv: the main idea of this aggregation strategy is to use an attention mechanism to learn which instructions are the most important for the classifier according to their position. In particular, we include in the network a vector $a$ that contains weights. The features $x_i$ of vertex $v_i$ is computed as a weighted mean of the vectors $\vec{\iota}$ of instructions $\iota \in I_{v_i}$. Note that the weights vector $a$ is trained end-to-end with the other network hyperparameters.  {\bf Rationale:} this strategy takes inspiration from recent works in neural machine translation \cite{bahdanau2014neural}. The presence of vector $a$ allows the network to automatically decide the importance of instructions relatively to their position inside each vertex.
\item \rnnTv: the feature $x_i$ of $v_i$ is the last vector of the sequence generated by a Recurrent Neural Network (RNN) that takes as input the sequence of vectors $\vec{\iota}$ of instructions $\iota \in I_{v_i}$. The RNN we consider in this work is based on GRU cell \cite{chung2014empirical}. This mechanism is trainable, the weights of the RNN are updated with the training of \sv. {\bf Rationale:} this method allows generate a vector representation that takes into account the order of the instructions in the input sequence. 
\end{itemize}

\subsection{Instruction Execution and Result Sampling (\sampTv)} 
We represent each block as a set of multivariable functions, and then compute features as sampling of these functions over random inputs.  This mechanism is not trainable.  {\bf Rationale:} this method is based on the assumption, common to \cite{crossarch}, that sampling a sequence of instructions captures a semantic that cannot be captured by static analysis.

\section{Graph Embedding Neural Network}\label{sec:theodetails}

\begin{figure*}[h]
\includegraphics[width=\textwidth]{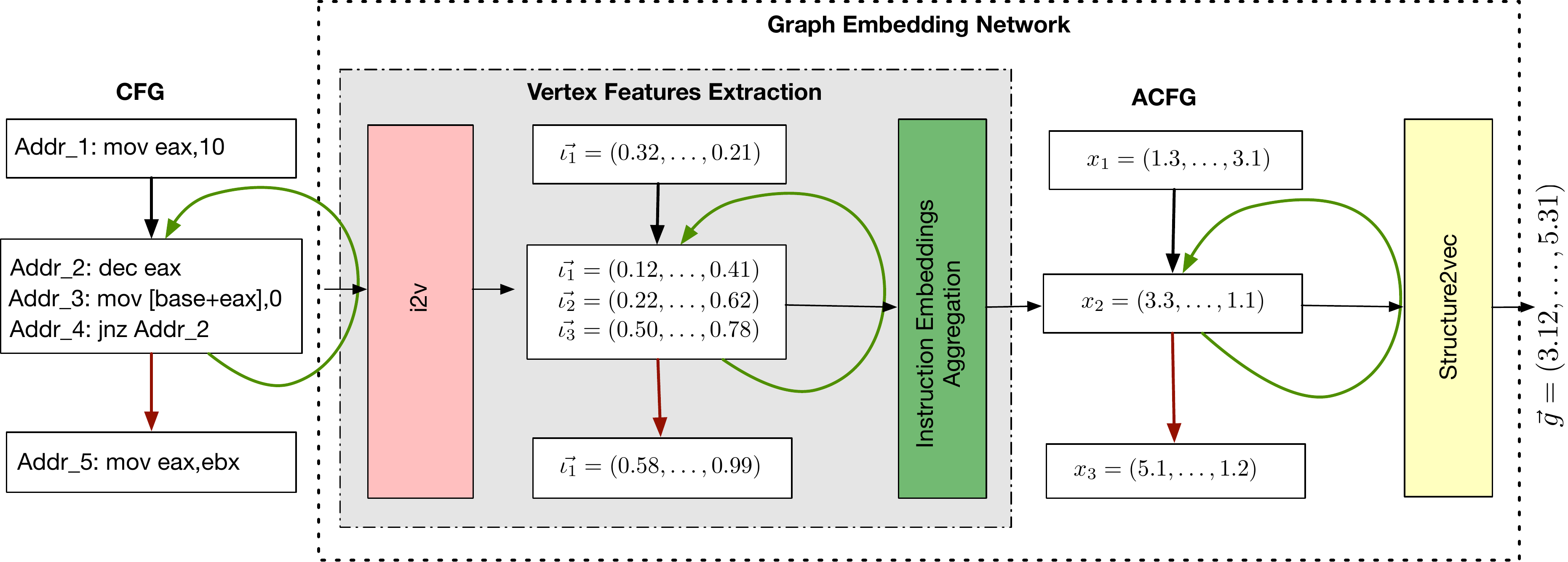}
\caption{Graph Embedding Neural Network Architecture. The vertex feature extractor component refers to the Unsupervised Feature Learning case. \label{fig:magg}}
\end{figure*}
We denote the entire network that compute the embedding of a CFG graph as {\em graph embedding neural network}.
The graph embedding neural network is the union of two main components: (1) the \emph{Vertex Feature Extractor}, that is responsible to associate a feature vector with each vertex in $g$, and (2) the \sv network, that combines such feature vectors through a deep neural architecture to generate the final embedding vector of $g$.
See Figure \ref{fig:magg} for a schematic representation of the overall architecture of the graph embedding network, where the  vertex feature extractor  component refer to the Unsupervised Feature Learning implementation.

\subsection{Vertex Features Extractor}
\label{sec:nfe}

The vertex feature extractor is the component where we focus the attention of this paper and where we propose most of our contributions.
The goal of this component is to generate a vector $x_i$ from each vertex $v_i$ in the CFG $g$. 
We considered several solutions to implement this component. 
As baseline we consider the approach based on manual feature engineering proposed in \cite{CCS}.
Moreover, we investigated solutions based on unsupervised feature learning (or representation learning), borrowing models and ideas from the natural language processing community.
These techniques allow the network to automatically discover the representations needed for the feature vectors from raw data.
This replaces manual feature engineering and allows the network to both learn the features and use them to generate the final graph embedding.
We will show in Section \ref{sec:evaluation} that this approach leads to performance improvements of the overall graph embedding network.
Finally we considered a solution that executes the different part of the CFG $g$ using a synthetic input and samples the results to generate the feature vector $x_i$.

\noindent\underline{Manual Feature Engineering (\sTv)}

As in \cite{CCS} for each block we use the following features: 
\begin{enumerate}
\item Number of constants;
\item Number of strings;
\item Number of transfer instructions (e.g. \textit{MOV});
\item Number of calls;
\item Number of instructions;
\item Number of arithmetic instructions (e.g. \textit{ADD});
\item Vertex offsprings;
\item Vertex betweenness centrality;
\end{enumerate}
The first six features are related to the code of the block. 
Instead, the last two features depend on the graph hence they bring some information about the structure of the control flow graph inside each vertex.

\noindent\underline{Unsupervised Feature Learning}

This family of techniques aim at discovering low-dimensional features that capture the underline structure of the input data.
The first step of these solutions consist in associating an embedding vector with each instruction $\iota$ contained in $I_{v_i}$.
In order to achieve this we train an embedding model \textsf{i2v} using the skip-gram method outlined in the paper that introduces word2vec technique for computing word embeddings \cite{Mikolov}. The idea of the skip-gram model is to use the current instruction to predict the instructions around it. A similar approach has beed used also in \cite{ChuaSSL17}.

We use as tokens to train the \textsf{i2v} model the mnemonics and the operands of each assembly instruction. 
Note that, we filter the operands of each instruction and we replace all base memory addresses with the special symbol \texttt{ MEM} and all immediates whose absolute value is above some treshold (we use $5000$ in our experiments, see Section \ref{sec:implementation}) with the special symbol \texttt{IMM}. 
The motivation behind this choice is that we believe that using raw operands is of small benefit, as example the relative displacement given by a jump is useless  (e.g., instructions do not carry with them their memory address), and, on the contrary, it may decrease the quality of the embedding by artificially inflating the number of different instructions.
As example the instruction \texttt{mov EAX,}$6000$ becomes \texttt{mov EAX,IMM}, \texttt{mov EAX,}$[0$\texttt{x}$3435423]$  becomes \texttt{mov EAX,MEM},
while the instruction \texttt{mov EAX,}$[$\texttt{EBP}$-8]$ is not modified. Intuitively, the last instruction is accessing a stack variable different from
\texttt{mov EAX,}[\texttt{EBP}$-4$], and this information remains intact with our filtering.


After we obtain an embedding for each instruction we still need to aggregate such vectors in order to obtain a single feature vector $x_i$ to associate with vertex $v_i$.
In this paper, we investigate three different instruction embeddings aggregation techniques: \aTv, \wTv, \rnnTv.

\subsubsection{\aTv}
This solution consists in averaging together all the instruction vectors in the vertex. 
In particular, the feature vector $x_i$ is obtained as follows:
\begin{equation}
	x_i = \frac{ \sum_{j=1}^{m} \vec{\iota_{j}} }{ m }
\end{equation}
where $m$ refers to the number of instructions in vertex $v_i$ (i.e., $|I_{v_i}|$) and $\vec{\iota_{j}}$ is the embedding vector of the $j$-th instruction in vertex $v_i$.

\subsubsection{\wTv}
Note that the \aTv solution does not take into consideration the order of the instructions in the block. To mitigate this problem we included in the network architecture an end-to-end trained vector $a$ that associates a different weight with each instruction position.
The feature vector $x_i$ equation is modified as follows:
\begin{equation}
	x_i = \frac{ \sum_{j=1}^{m} a[j] \cdot \vec{\iota_{j}} }{ \sum_{j=1}^{m} a[j] }
\end{equation}
where $a[j]$ is the $j$-th component of vector $a$.

\subsubsection{\rnnTv}
To fully take into consideration the instruction position we also considered a solution that incorporates a Recurrent Neural Network (RNN) \cite{hochreiter1997long} into the overall network architecture. 
This RNN is trained end-to-end, takes in input all the instruction embedding vectors in order, i.e.  $( \vec{\iota_1}, ..., \vec{\iota_m} )$ and generates $m$ outputs and $m$ hidden states $( h^{(1)}, ..., h^{(m)} )$. The final feature vector $x_i$ is simply the last hidden state of the RNN, that is:
\begin{equation}
	x_i = h^{(m)}
\end{equation} 

%

\noindent\underline{Instruction Execution and Result Sampling (\sampTv)}

Given a vertex of the CFG, we partition it in sequences of instructions that update
independent memory locations, these are the {\em strands} defined in \cite{David}.
We construct an extended version of the CFG that we call SCFG.
The SCFG contains all strands of the CFG as vertices and its topology is built as follows: 
for each vertex $v_i$ of the CFG we define a total order on the vertex's strands. We take all predecessors of vertex $v_i$ in the CFG
and in the SCFG we add an edge from each predecessor's last strand to the first strand in $v_i$. Finally, we create an oriented path connecting all strands in $v_i$ according to their order.


To transform a strand in a set of functions we execute it symbolically, using the ANGR framework.
Obtaining multivariable functions expressed as formulas for the Z3 solver \cite{Z3}:
Each function defines the value of a memory location that is wrote without being read after (output), as combination of 
a set of inputs (an input is a memory location that is read before any write on it and that concurs to the value of the output).

Finally, we compute a feature vector from each strand by sampling and averaging the outputs of the strand's functions over 100 randoms inputs.

Note that functions are in general non symmetric; This can create a problem since two semantically equivalent strands may have different ordering of their inputs. 
We address this problem by symmetrising the function: 

Given a function $z: \mathbb{R}^n \rightarrow \mathbb{R} $ we define its symmetrisation as:
\begin{equation}
\label{eq:simmetrization}
z'(\vec{b}) =\frac{1}{|\Pi(\vec{b})|} \sum_{\vec{b'} \in \Pi(\vec{b})} z(\vec{b'})
\end{equation}

where $\Pi(\vec{b})$ is the set of all the possible vectors obtained by permuting the components of vector $\vec{b}$. The above implies that we have to compute
$z$ on each permutation of its inputs. This is a costly operation, the subdivision in strands it is beneficial: subdividing a vertex in smaller units reduces the number of variables 
in each of the segments that are symbolically executed. 
Unfortunately, it is not enough.  Therefore, cap the number of inputs to $5$ (this
threshold includes almost all functions in our test dataset). More precisely, we order the inputs in an arbitrary way and we forces all inputs from position 6 on to value zero. 
Combining symmetrisation and the average of all functions outputs we obtain a feature array of size 100 for each vertex of the SCFG.

\begin{figure*}[t!]
\centering
\subfloat[Vertex-specific $\mu_{v}$ vectors are updated according to graph topology and vertex features.]{
\includegraphics[width=.46\textwidth]{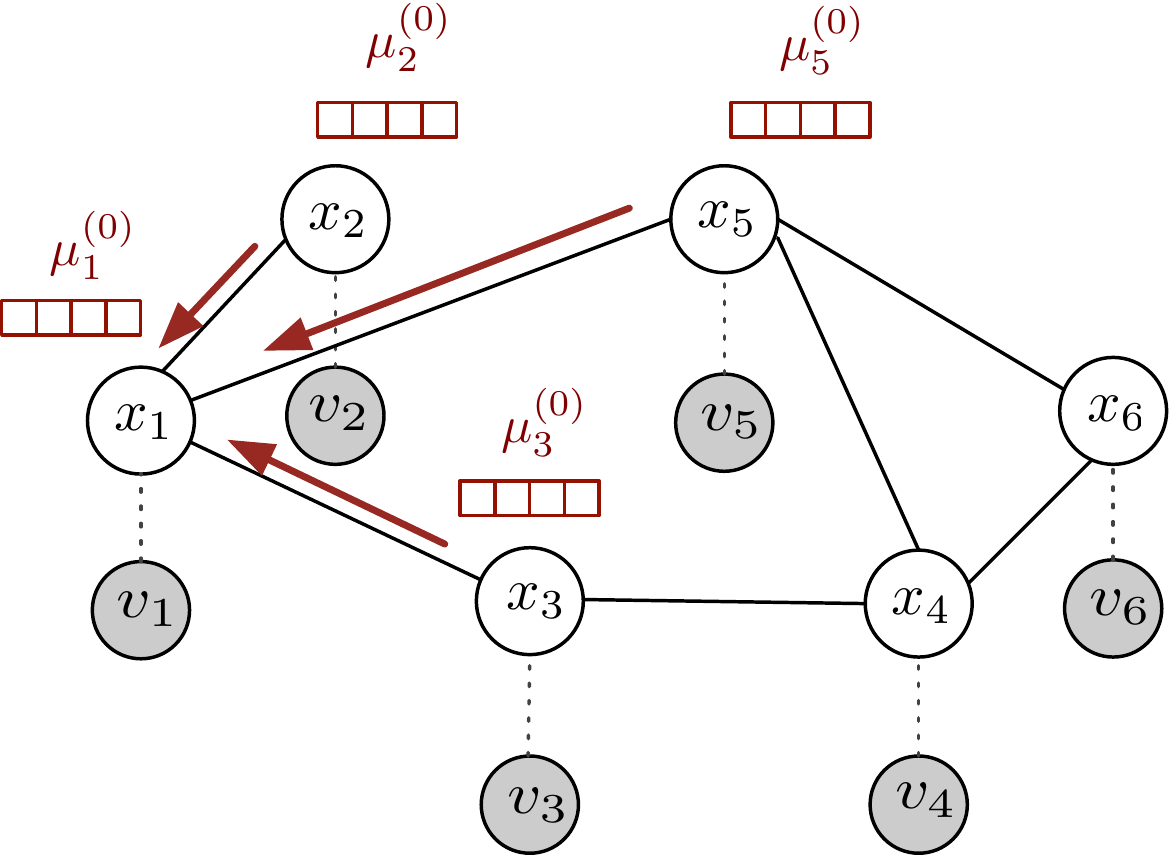}}\hfill
\subfloat[The $\mu$ vectors update process includes an $\ell$ layer fully-connected neural network with  $\text{ReLU}$ activation.]
{\includegraphics[width=.46\textwidth]{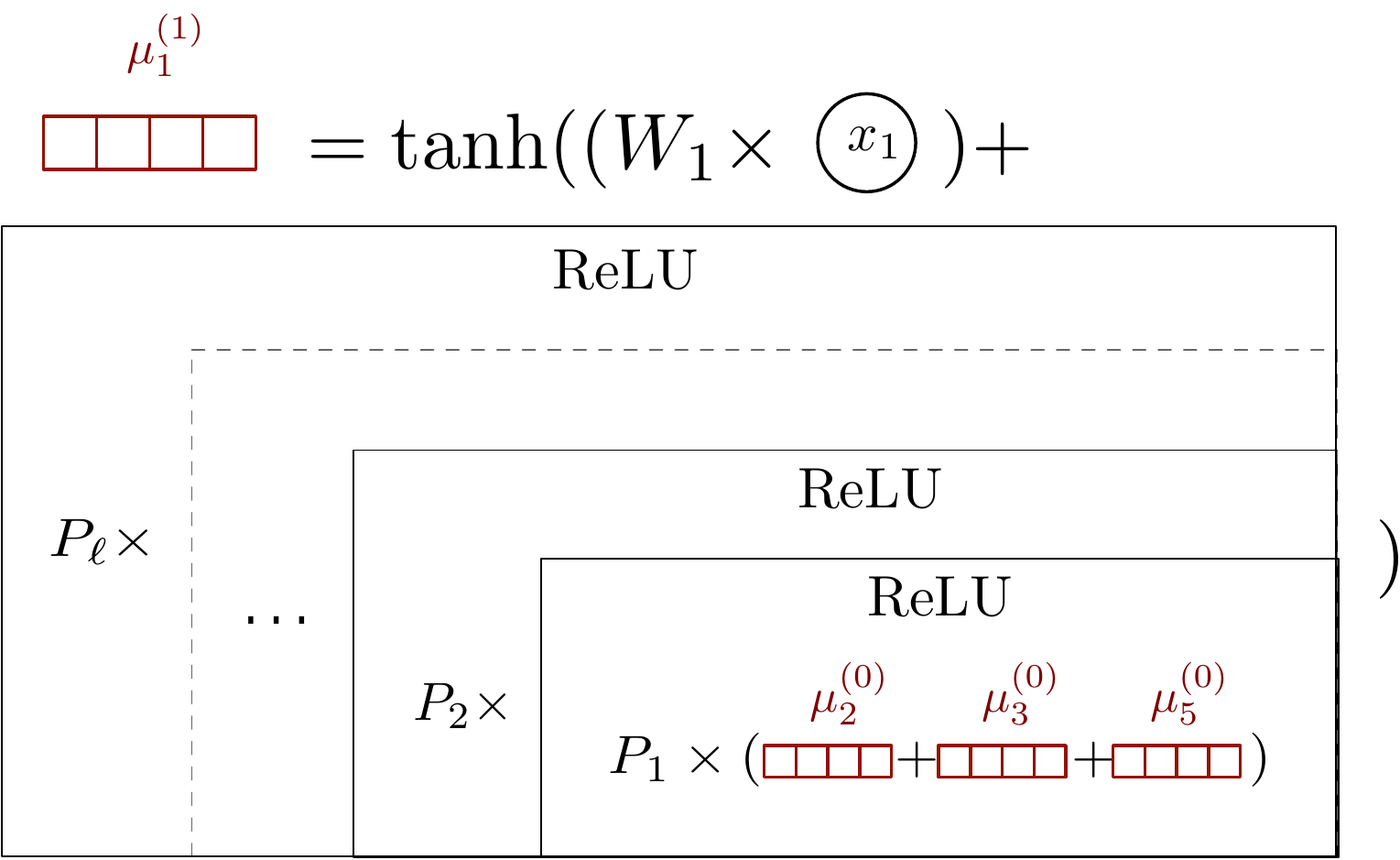}}\hfill
\subfloat[The final graph embedding $\vec{g}$ is obtained by aggregating together the vertex $\mu$ vectors after $T$ rounds.]
{\includegraphics[width=.46\textwidth]{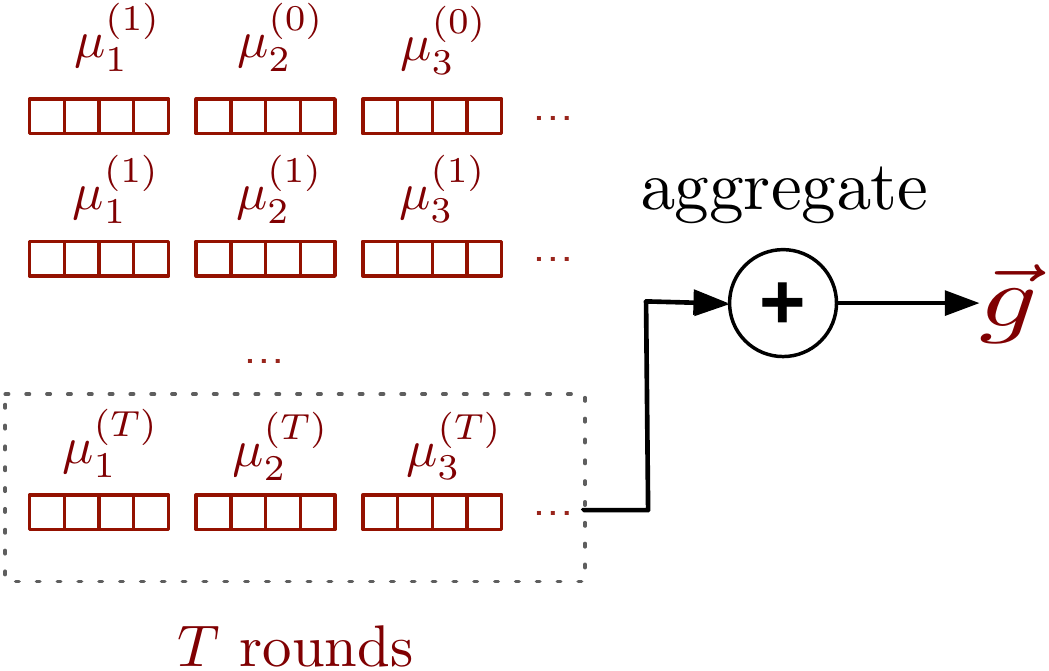}}\hfill
\subfloat[Siamese Architecture that uses two identical Graph Embedding Networks and a similarity score to learn the network parameters.]
{\includegraphics[width=.46\textwidth]{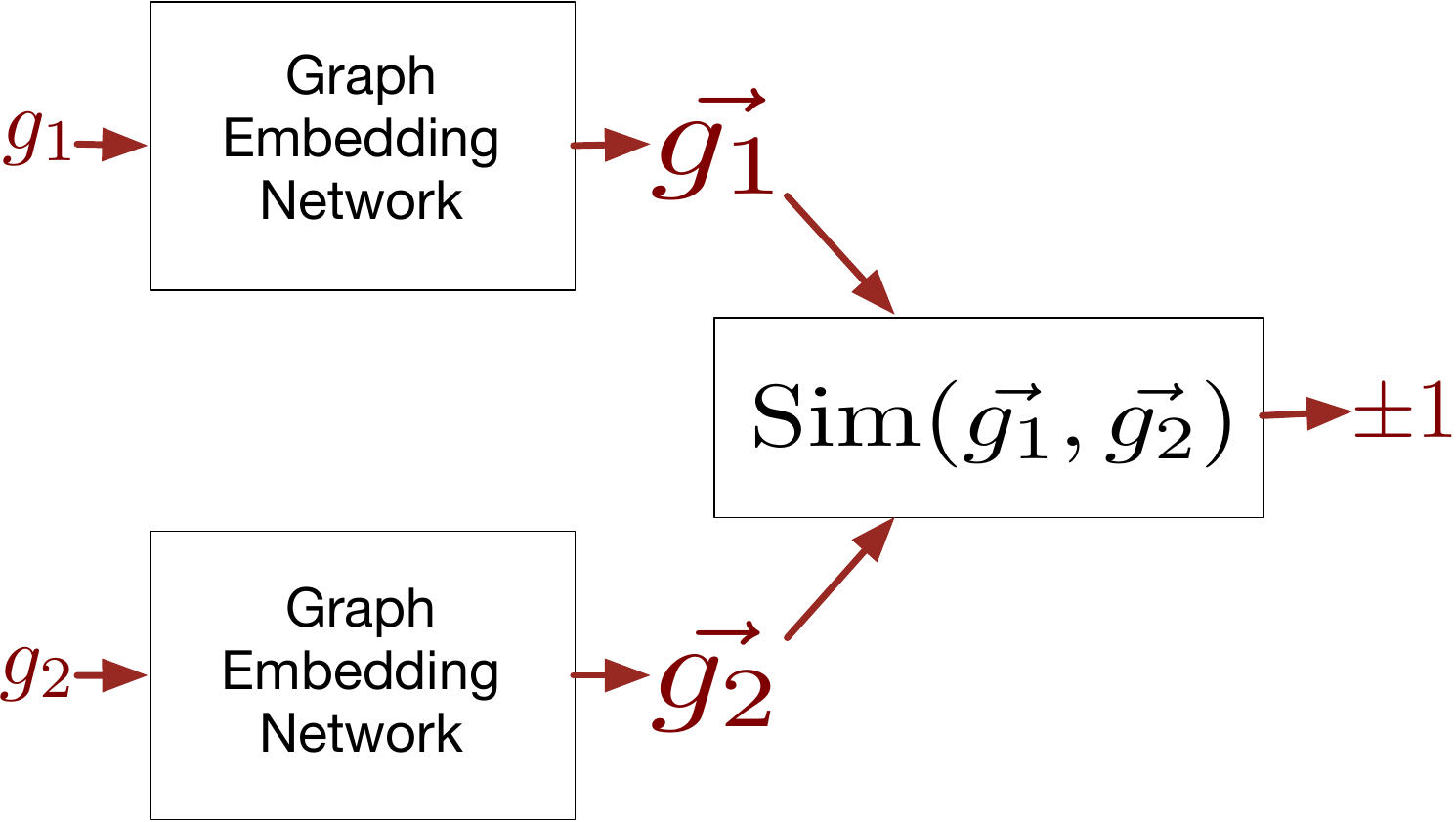} \label{fig:siamese}}
\caption{Structure2Vec Deep Neural Network and Siamese Architecture to learn parameters.}
\label{fig:fig_structure2vec}
\end{figure*}

\subsection{Structure2Vec Deep Neural Network} \label{sec:ne}

The \sv component is based on the approach of \cite{dai2016discriminative} using the parameterization of \cite{CCS}.
In order to compute the embedding of the graph $g$, a $p$-dimensional vector $\mu_{i}$ is associated with each vertex $v_i$.
The $\mu$ vectors are dynamically updated during the network training following a synchronous approach based on rounds.
We refer with $\mu_{i}^{(t)}$ to the $\mu$ vector associated with vertex $v_{i}$ at round $t$.

We aggregate vertex-specific $\mu$ vectors and features following the topology of the input graph $g$.
After each step, the network generates a new $\mu$ vector for each vertex in the graph taking into account both the vertex features and graph-specific characteristics.
In particular, the vertex vector $\mu_{i}$  is updated at each round as follows:
\begin{equation}
	\mu_{i}^{(t+1)} =  \mathcal{F} \Big( x_{v_i}, \sum_{u_{j} \in \mathcal{N}(v_i)} \mu_{j}^{(t)}  \Big),   \forall v_i \in V
\end{equation} 

The vertex $\mu$ vectors at round zero $\mu^{(0)}$ are randomly initialized and $\mathcal{F}$ is a nonlinear function:
\begin{equation}
	\label{eq:f}
	\mathcal{F} \Big( x_{v_i}, \sum_{u_j \in \mathcal{N}(v_i)} \mu_j^{(t)}  \Big) = \text{tanh} \Big( W_1 x_{v_i} + \sigma (   \sum_{u_j \in \mathcal{N}(v_i)} \mu_{j}^{(t)}  ) \Big)
\end{equation}
where $W_1$ is a $d \times p$ matrix, $tanh$ indicates the hyperbolic tangent function and $\sigma$ is a nonlinear function:
\begin{equation}
	\label{eq:sigma}
	\sigma(y) = P_1 \times \text{ReLU} ( P_2 \times ...\text{ReLU}(P_\ell \times y))
\end{equation}

The function $\sigma(y)$ is an $\ell$ layers fully-connected neural network, parametrized by $\ell$ matrices $P_i (i = 1, ..., \ell)$ of dimension $p \times p$.
$\text{ReLU}$ indicates the rectified linear unit function, i.e., $\text{ReLU}(x) = max\{0,x\}$.

The final graph embedding $\vec{g}$ is obtained by aggregating together the vertex $\mu$ vectors after $T$ rounds, as follows:
\begin{equation}
	\vec{g} = W_2 \sum_{v_i \in V} \mu_i^{(T)} 
\end{equation}
where W$_2$ is another $p \times p$ matrix used to transform the final graph embedding vector.

\noindent\underline{Learning Parameters Using Siamese Architecture}

To learn the network parameters $\Phi = \{ W_1, W_2, P_1, ..., P_\ell\}$ we use, as in \cite{CCS}, a pairwise approach, a technique also called \emph{siamese network} in the literature \cite{bromley1994signature}.
The main idea is to use two identical graph embedding networks - i.e., the two networks share all the parameters - and join them with a similarity score.
That is, the final output of the siamese architecture is the similarity score between the two input graphs.
In particular, from a pair of input graphs $< g_1, g_2 >$ two vectors $< \vec{g_1}, \vec{g_2} >$ are obtained by using the same graph embedding network.
These vectors are compared using cosine similarity as distance metric, with the following formula:
\begin{equation}
	\text{similarity}(\vec{g_1}, \vec{g_2}) = \frac{ \displaystyle\sum_{i=1}^{p} \Big( \vec{g_1}[i] \cdot \vec{g_2}[i]  \Big) }{  \sqrt{  \displaystyle\sum_{i=1}^{p}  \vec{g_1}[i]  } \cdot \sqrt{  \displaystyle\sum_{i=1}^{p}  \vec{g_2}[i]  }  }
\end{equation}
where $\vec{g}[i]$ indicates the $i$-th component of the vector $\vec{g}$.
The overall siamese architecture is illustrated in Figure \ref{fig:siamese}.

To train the network we require in input a set of $K$ CFGs pairs, $< \vec{g_1}, \vec{g_2} >$, with ground truth labels $y_i \in \{+1,-1\}$, where $y_i = +1$ indicates that the two input graphs are similar and $y_i = -1$ otherwise. Then we compute the siamese network output similarity for each pair and we define the following least squares objective function:
\begin{equation}
	J = \displaystyle\sum_{i=1}^{K} \Big(  \text{similarity}(\vec{g_1}, \vec{g_2}) - y_{i}  \Big)^2
\end{equation}

The objective function $J$ is minimized by using, for instance, stochastic gradient descent.
Note that the vertex feature extractor component might add additional hyperparameters to $\Phi$ (as, for instance, the RNN hyperparameters for  \rnnTv). 
All these parameters are jointly learned end-to-end by minimizing $J$.


\section{Evaluation}\label{sec:evaluation}

We conducted an experimental study on datasets extracted from real binaries collections to compare our vertex feature extraction solutions with other state-of-the-art approaches.

\subsection{Datasets}
\label{sec:dataset}

We constructed three datasets: \normalDataset, that is used as a benchmark to compare our vertex features against baseline state-of-the-art approaches; two large copora of assembly source codes used to train two \textsf{i2v} models, one for the AMD64 instruction set and one for the ARM instruction set; and a \simDataset that is used to do a qualitative analysis of the function embeddings. 


\subsubsection{\normalDataset}
To align our experimental evaluation with state-of-the-art studies we built the \normalDataset in the same way as the one used in \cite{CCS}. In particular, the \normalDataset consists of a set of 95535 graphs generated from all the binaries included in two versions of Openssl (v1\_0\_1f - v1\_0\_1u) that have been compiled for X86 and ARM using gcc-5.4 with 4 optimizations levels (i.e., -O$[0$-$3]$). The resulting binaries have been disassembled using ANGR\footnote{ANGR is a framework for static and symbolic analysis of binaries} \cite{angr} and we discarded all the functions that ANGR was not able to disassemble.

\subsubsection{Assembly source codes corpus to train \textsf{i2v}}
\label{dataset:i2v}
We built two different \textsf{i2v} models, one for the AMD64 instruction set and one for the ARM instruction set.
With this choice we tried to capture the different syntactic and semantics of these two assembly languages.
We collected the assembly source code of a large number of assembly functions to build two training corpora, one for the \textsf{i2v} AMD64 model and one for the \textsf{i2v} ARM model.
We built both corpora by dissasembling several unix executables and libraries using IDA PRO\footnote{We used IDA PRO because of its performance in disassembling executables. However, we built our entire prototype on the open source framework ANGR because we intend to release an open source fully functional prototype.}. 
We avoided the multiple inclusion of common functions and libraries by using a duplicate detection mechanism; we tested the uniqueness of a function computing an hash of all function instructions, where instructions are filtered by replacing the operands containing immediate and memory locations with a special symbol.
From 2.52 GBs of AMD64 binaries we obtained the assembly code of 547K unique functions. 
From 3.05 GBs of ARM binaries we obtained the assembly code of 752K unique functions.
Overall the AMD64 corpus contains $86$ millions assembly code lines while the ARM corpus contains $104$ millions assembly code lines.


\subsubsection{\simDataset for qualitative analysis of function embeddings} the \simDataset has been generated from source codes collection of 410 functions that have been manually annotated as implementing algorithms of encryption, sorting, string operations (e.g., string reversing, string copy) and statistical analysis (e,g, average, standard deviation, ecc). Note that \simDataset contains multiple functions that refer to different implementations of the same algorithm.  We compiled the sources for AMD64 using clang-3.9 and gcc-5.4 with -O$[0$-$3]$, and after disassembling and removing duplicated functions we obtained from the 410 functions a total of 2716 CFGs.


\subsection{Implementation Details} \label{sec:implementation}

We developed a prototype implementation of the graph embedding neural network and of all the different vertex feature extraction solutions using Python and the Tensorflow \cite{abadi2016tensorflow} framework\footnote{The source code of our prototype as well as a virtual machine containing the datasets and the scripts to reproduce our results will be made publicly available. We did not published the prototype to not give away the anonymity of the reviewing process. It will be published once we are not anymore bounded to anonymity constraint.}.
For static analysis and symbolic execution of binaries we used the ANGR framework \cite{angr}.  
To train the network we used a batch size of 250, learning rate $0.001$, Adam optimizer, feature vector $|x_i|=100$, function embeddings of dimension $p=64$ (this is the same dimension of $\mu$ vectors), number of rounds $T=2$, and a number of layers in  Structure2vec $\ell=2$. 
These values have been chosed with an exhaustive grid search over the hyperparameters space.
Tensorflow requires training batches of uniform dimension, therefore we manipulate each vertex $v_i$ to contain the same number of instructions; i.e. we fix the length of $I_{v_i}$, this is done by either padding with a special instruction or by truncation. 
The padding vectors contain all zeros.
We decided to fix the number of instructions inside each vertex to $m=150$ (all vertices in all graphs of our dataset contain less than $150$ instructions). 
To keep the training batches uniform we also decided to pad the number of vertices in each CFG to $100$, removing the CFGs above this threshold. 
In our dataset less than 4\% of the graphs are above such threshold. 
Finally, the RNN used in \rnnTv is a multi-layer network with $3$ layers and GRU cell.


We trained two \textsf{i2v} models using the two training corpora described in Section \ref{dataset:i2v}, one for the instruction set of ARM and one for AMD64. 
The model that we use for \textsf{i2v} (for both versions AMD64 and ARM) is the skip-gram implementation of word2vec provided in \cite{w2vimp}.
We used as parameters: embedding size 100, window size $8$ and word frequency $8$. 
To validate the benefits of an instruction embedding model we experimented also with random embeddings. In particular, we associate a random vector to each instruction appearing more than $8$ times in the training documents described in Section \ref{dataset:i2v}.
All instructions appearing less than $8$ times are mapped into the same  random vector.

%
%
%
%


\subsection{Test Methodology}

We designed the methodology used in our tests following the one in \cite{CCS}. More precisely, we generate our training and test pairs as reported in \cite{CCS}, and we use as test measure the area under the receiver operating characteristic curve of the classifiers. Moreover, we additionaly use a $5$-fold cross validation to compare the performances of our best model and the one in \cite{CCS}. Details follow. 

In order to train and test our system, we create a certain number of pairs using the \normalDataset. The pairs can be of two kinds: similar pairs, obtained pairing together two CGFs originated by the same source code, and dissimilar pairs, obtained pairing randomly CFGs that do not derive from the same source code. 
Specifically, for each CFG in our dataset we create two pairs, a similar pair, associated with training label $+1$ and a dissimilar pair, training label $-1$; obtaining a total number of pairs $K$ that is twice the total number of CFGs. 
We split these pairs in three sets: train, validation,  and test.  As in \cite{CCS} pairs are partitioned preventing that two similar CFGs are in different sets (this is done to avoid that the network sees during training graphs similar to the ones on which it will be validated or tested).
The split is 80\%-10\%-10\%. 
We train our models for $50$ epochs (an epoch represents a complete pass over the whole training set) and we compute performance metrics on the validation set for all the epochs.
Then, we use the model hyperparameters that led to the best performance on the validation set to compute a final performance score on the test set.
In each epoch we regenerate the training pairs, that is we create new similar and dissimilar pairs using the graphs contained in the training split. We precompute the pairs used in each epoch, in such a way that each method is tested on the same data. 
Note that, we do not regenerate the validation and test pairs. 

We used the static train/validation/test split in our first set of experiments to understand which vertex feature extraction model perform the best, then we performed an additional set of experiments comparing the best performing solution with the baseline approach using 5-folds cross validation. In the 5-folds cross validation we partitions the dataset in $5$ sets; for all possible set union of 4 partitions we train the classifiers on such union and then we test it on the remaining partition. 
The reported results are the average of 5 independent runs, one for each possible fold chosen as test set.
This approach is more robust than a fixed train/validation/test split since it reduces the variability of the results.



In general, we have four possible outcomes: either (i) correct predictions, both for similar pairs (TP true positives), and dissimilar pairs (TN true negatives)
or (ii) wrong predictions, both if $+1$ is predicted for an observed dissimilar pair (FP false positives) or if $-1$ is predicted for an observed similar pair (FN false negatives). These four values constitute the so called confusion matrix of the classifier.
Starting from the confusion matrix we can further define the algorithm \emph{sensitivity} (or True Positive Rate (TPR)), i.e. the ratio between TP and TP + FN, that measures the proportion of actual similar pairs which are correctly identified as such. In the same way we can define the algorithm \emph{specificity} (or True Negative Rate (TNR)), i.e. the ratio between TN and TN + FP, that measures the proportion of dissimilar pairs which are correctly identified as such. Their complementary values are represented by the False Positive Rate (FPR), i.e. $\frac{1}{\text{TNR}}$, and False Negative Rate (FNR), i.e. $\frac{1}{\text{TPR}}$. 
The \emph{Receiver operating characteristic} (ROC) curve \cite{herlocker2004evaluating} visually illustrates the performance of a binary classifier by plotting different TPR (Y-axis) and FPR (X-axis) values. The diagonal line of this graph (the so-called line of no-discrimination) represents a completely random guess: classifiers represented by this line are no better than a random binary number generator. A perfect classifier would be represented by a line from the origin $(0,0)$ to the upper left $(0,1)$ plus another line from upper left $(0,1)$ to upper right $(1,1)$. Real classifiers are represented by curves lying in the space between the diagonal line and this ideal case.
The area under the ROC curve, or AUC (Area Under Curve), is a popular evaluation metric that measures the two-dimensional area underneath the entire ROC curve. Higher the AUC value, better the predictive performance of the algorithm. 
\subsection{Results}

{\renewcommand{\arraystretch}{1.25}%
\begin{table*}[t]

	\centering
	\small
	\begin{tabular}{ c | c | c | c | c  | c   }

		Index     	& Vertex Feature Extractor	& \multicolumn{2}{c |}{Instruction Representation}		  	& Val Auc 			& Test Auc 					\\ 
		\hline
		1					& \sTv  	& \multicolumn{2}{c|}{-}							&	\textit{0.946}		&	\textit{0.950}	  	\\	
		\cline{2-6}
		2					& \multirow{4}{*}{\aTv}   	&  \multirow{2}{*}{\textsf{i2v}}	&	Static &	{0.956}		&	{0.954}	  	\\	
		\cline{4-6}
		3					&   	& 		&	Non-Static							&	0.908					&	0.909	  				\\	
		\cline{3-6}
		4					&   	& \multirow{2}{*}{Random Embedding} 			&	Static							&	0.924					&	0.918	  				\\	
		\cline{4-6}
		5					&   	& 			&	Non-Static							&	0.893					&	0.891	  				\\	
		
		\cline{2-6}
		6					& \multirow{4}{*}{\wTv}  	& \multirow{2}{*}{\textsf{i2v}}	&	Static							&	\textbf{0.961}		&	\textbf{0.956}	  	\\	
		\cline{4-6}
		7					&   	& 		&	Non-Static							&	0.926					&	0.917	  				\\
		\cline{3-6}
		8					&   	& \multirow{2}{*}{Random Embedding} 	 		&	Static							&	0.887					&	0.881	  				\\	
		\cline{4-6}
		9					&   	& 		&	Non-Static							&	0.933					&	0.932	  				\\	
		\cline{2-6}
		10					& \multirow{1}{*}{\rnnTv}  	& \multirow{1}{*}{\textsf{i2v}}		&	Static							&	{{0.847}}				&	{{0.843}}  		\\	
		\cline{2-6}
		14					& \multirow{1}{*}{\sampTv}  	& 	 \multicolumn{2}{c|}{-}								&	 0.677					&	0.710	  				\\	
	\end{tabular} 
	\vspace{0.5cm}
		\caption{Test over \normalDataset. In italic we report the results obtained reproducing \cite{CCS}. In bold we report our best results. All the experiments were conducted on the \normalDataset.}
	\label{table:1}
\end{table*}


We present the results of two sets of experiments where we compare the considered vertex features extraction solutions on the \normalDataset.

Table \ref{table:1} shows the results of our first set of experiments, conducted on the fixed train/test/validation split of the \normalDataset. The entries in the last two columns shows the best AUC value obtained on the validation set and the corresponding AUC value for the test set. 
First note that the best performing vertex feature extraction solution is \wTv with static pre-trained \textsf{i2v} embedding (i.e., index 6 - $0.956$ AUC on test set), followed by \aTv with static pre-trained \textsf{i2v} embedding (i.e., index 2 - $0.954$ AUC on test set). This highlights the benefits of adopting an unsupervised approach to feature learning with respect to a manual feature selection approach. Moreover, both these solution benefit from the pre-trained \textsf{i2v} models. In fact, pre-trained \textsf{i2v} vectors led to a boost in performance ranging from $4\%$ to $9\%$ with respect to randomly chosen instruction vectors. 
Note that we performed experiments both keeping instruction embedding vectors static throughout training (i.e., \emph{static} label) or allowing the network to update them via backpropagation (i.e., \emph{non-static} label). The best performance for both \wTv and \aTv are obtained with static instruction embedding vectors. A possible explanation for this behaviour is that keeping these vectors static allows a higher flow of information from structure learnt in the big corpus used to pre-train the \textsf{i2v} model and the overall graph embedding network - i.e., the network does not force these embedding to fit/overfit the specific \normalDataset.
A similar phenomena often happens also in the context of natural language processing, see for instance \cite{kim2014convolutional}. 
The \rnnTv solution performs quite poorly. This is probably due to the high number of parameters that the network has to train, the considered dataset might contain not enough training points to accurately train all of the parameters or simply the network is too complicated for this specific task.
The \sampTv solution performs particularly poorly, achieving an AUC score on the test set of 0.710. The reason behind this poor performance is probably due to increased complexity of the SCFG graph (recall that SCFG represents the program flow among strands, and this leads to a graph that has more edges and a more intricate structure that the CFG, see Section \ref{sec:theodetails}), and/or to the poor quality of the features obtained by sampling. However, a further investigation is needed to confirm that such sampling strategy is indeed a failure.  


\begin{figure}[!t]
\includegraphics[width=\linewidth]{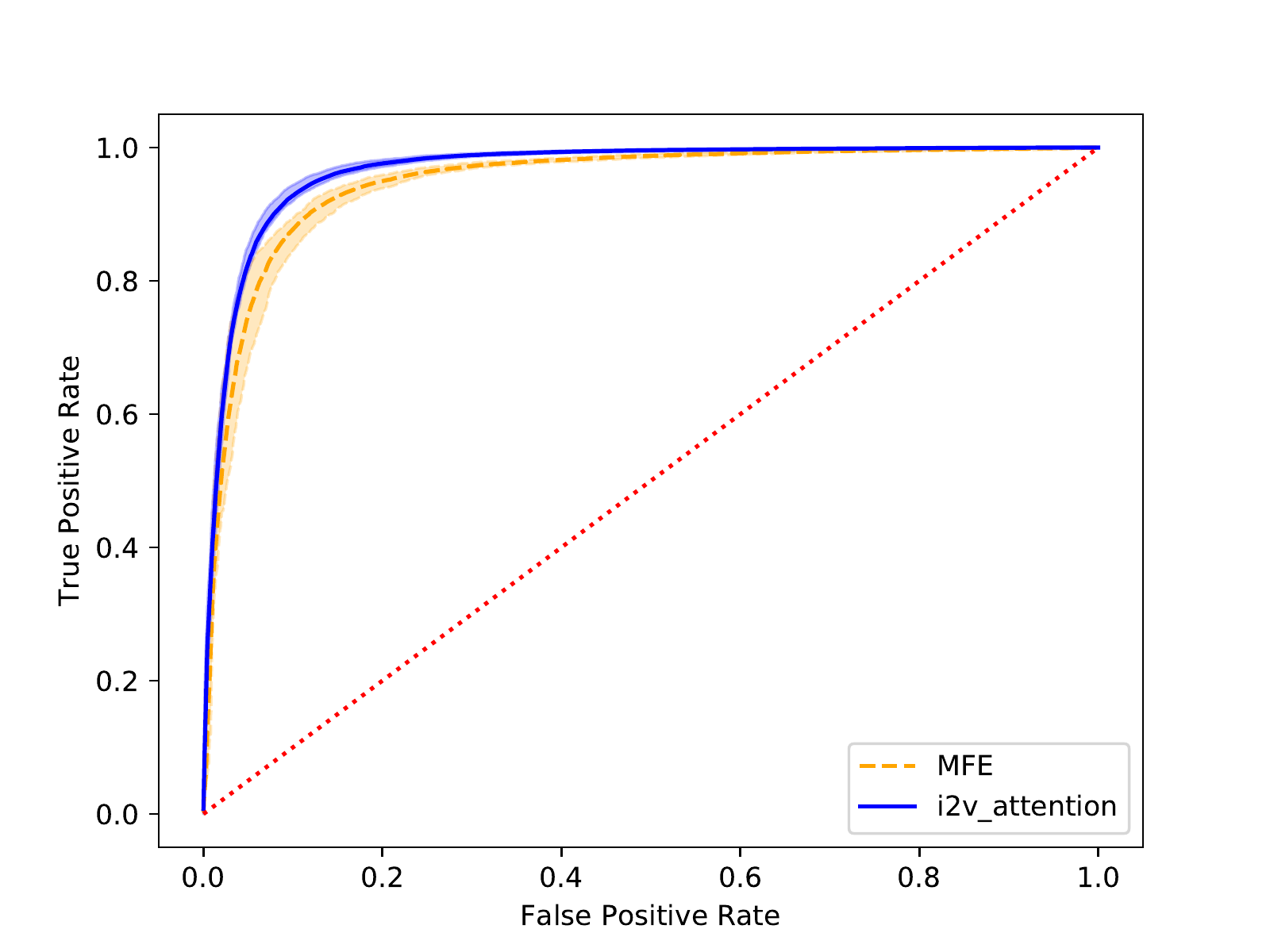}
\caption{ROC curves for the comparison between \sTv and \wTv with static pre-trained \textsf{i2v} embedding, using 5-fold cross validation. The lines represent the ROC curves obtained by averaging the results of the five runs; the dashed line is the average for \sTv, the continuous line the average for \wTv. For both \sTv and \wTv we color the area between the ROC curves with minimum AUC and the maximum AUC.}\label{fig:roc_curve}
\end{figure}


Stimulated by the result obtained in the first set of experiments we decided to perform a second set of experiments where we compare \wTv with static pre-trained \textsf{i2v} embedding and \sTv in a more robust way. 
This is to further confirm our hypothesis of the superiority of an automatic unsupervised feature learning approach with respect to manual feature engineering. 
In particular, we performed a comparison using a $5$-fold cross validation approach on \normalDataset. 
Figure \ref{fig:roc_curve} shows the average ROC curves of the five runs. The \sTv results are reported with an orange dashed line while we used a continuous blue line for the \wTv results. For both solutions we additional highlighted the area between the ROC curves with minimum AUC maximum AUC in the five runs.
The better prediction performance of the \wTv solution are clearly visible; the average AUC obtained by \sTv is $0,948$ with a standard deviation of $0,006$ over the five runs, while the average AUC of \wTv is $0,964$ with a standard deviation of $0,002$.  
Overall we observed an improvement of almost $2\%$ of \wTv with respect to \sTv. This clearly confirm the benefit of learning vertex features with respect to manually engineering them.
Note also that the standard deviation of the AUC values over the five runs is smaller for \wTv than for \sTv.
A final remark on the comparison; \cite{CCS} states that removing the betweenness centrality from the features used by \sTv slightly improves the performance of \sTv. We tried to confirm this but we found no improvement on the normal \sTv. 

\subsection{Qualitative Embeddings Evaluation}

We perform two qualitative analyses on the embedding vector space, considering both the overall binary function embeddings and the inner assembly instruction embeddings.


\begin{figure*}[t!]
\centering
\subfloat[2-dimensional visualisation of the embedding vectors for the 2716 binary functions in \simDataset. The four different categories of algorithms (encryption, sorting, statistical and string) are represented with different symbols and colors.]
{\includegraphics[width=.48\textwidth]{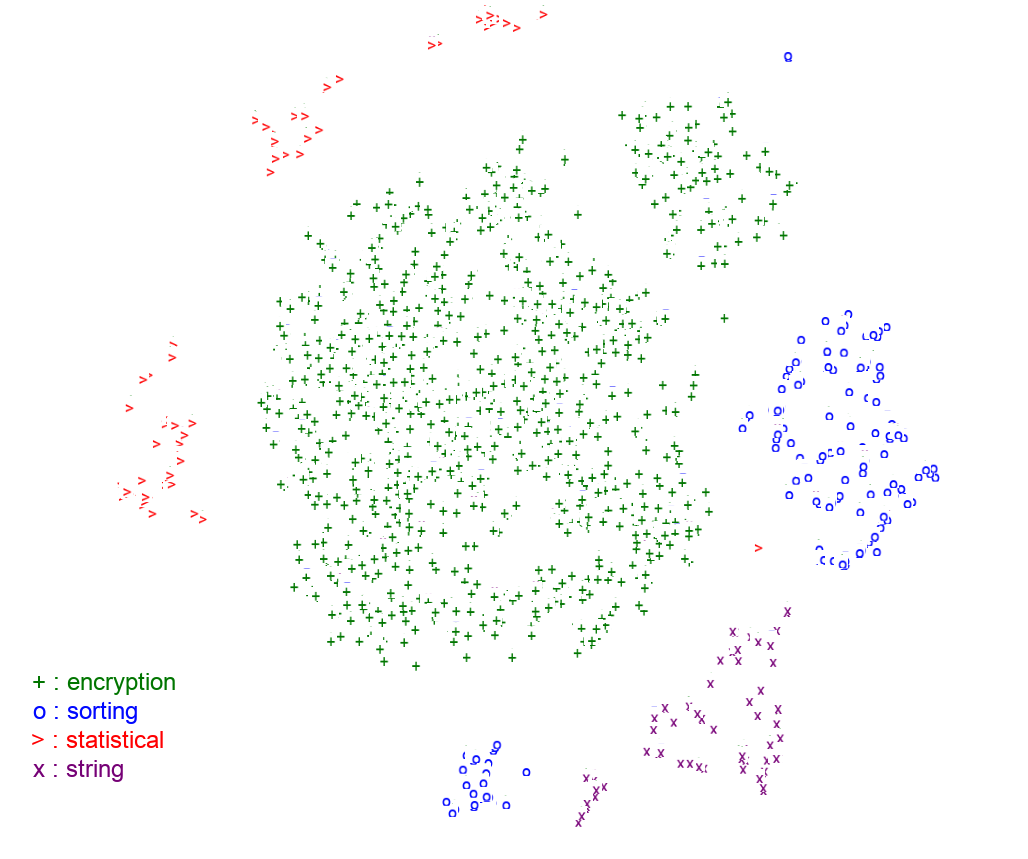} \label{sf:2}}\hfill
\subfloat[A zoom on the portion of the vector space containing binary function implementing encryption algorithms. We highlighted with a triangle, a circle and a square shape three interesting findings: implementations of an AES helper function, different optimisations of the Camellia encryption and decryption algorithm respectively.]{
\includegraphics[width=.48\textwidth]{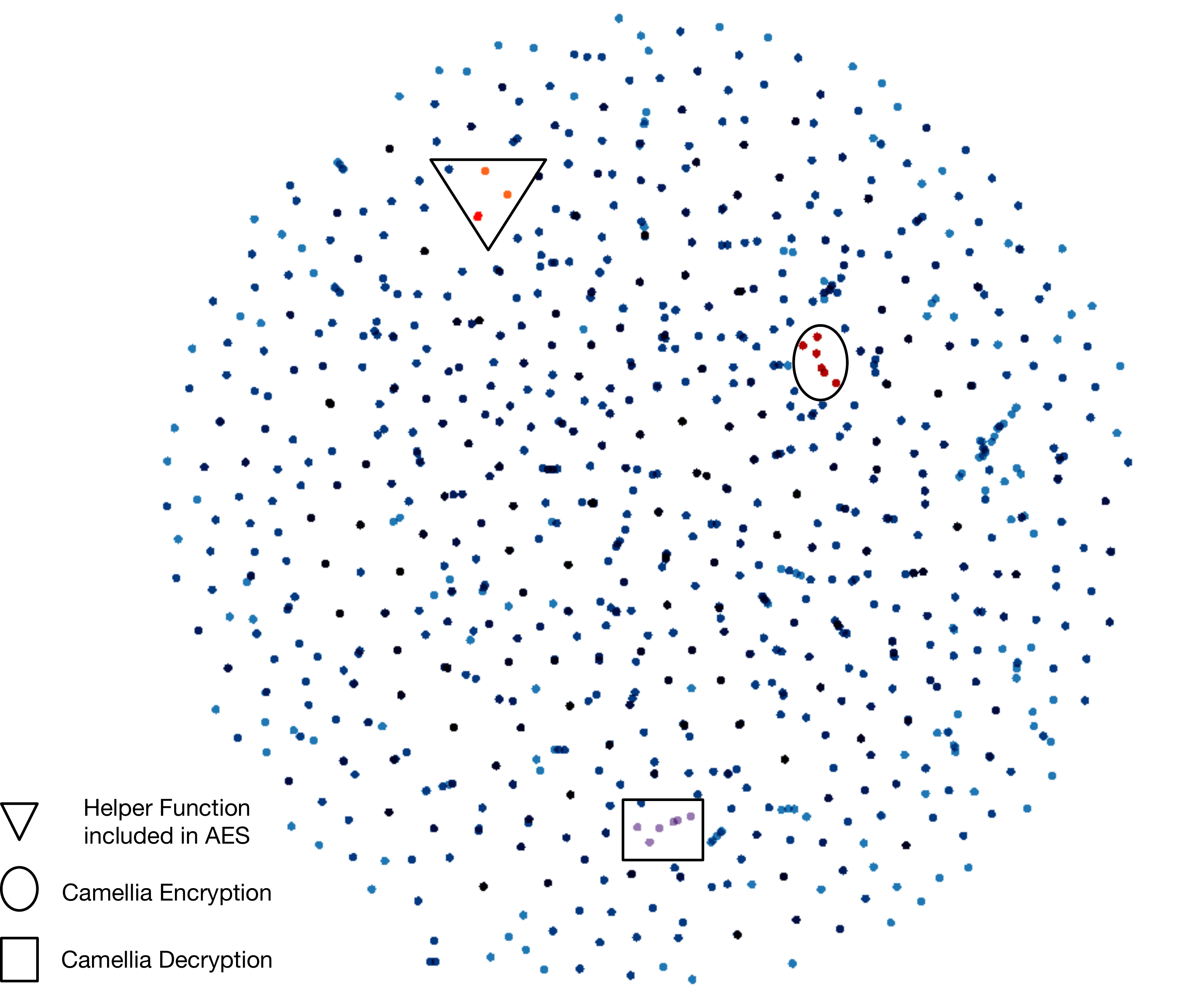} \label{sf:1}}\hfill
\caption{Qualitative analysis of embeddings vectors for binary functions. The points have been clustered using t-SNE.}
\label{fig:functions_embeddings}
\end{figure*}


The main objective of the first qualitative analysis we performed was to understand if the graph embedding network was able to capture information on the inner semantics of the binary functions, and to represent such information in the vector space.
To this end we used the \simDataset, a collection of 2716 manually annotated binary functions implementing one of the following four typologies of algorithms:   encryption, sorting, string operations and statistical analysis.
In particular, we used the best performing graph embedding network trained on \normalDataset, that is \wTv with static pre-trained \textsf{i2v} embedding, to generate an embedding vector for all of the considered 2716 binary functions.
Figure \ref{fig:functions_embeddings} shows a two-dimensional projection of the $64$-dimensional vector space where binary functions embeddings lie, obtained using the \emph{t-SNE}\footnote{We used the TensorBoard implementation of \emph{t-SNE}} visualisation technique \cite{maaten2008visualizing}.
From Figure \ref{sf:2} is possible to observe a quite clear separation between the different typologies of algorithms considered. We believe this behaviour is really interesting, it seems to suggest that embeddings computed with our technique could be used to classify the semantic of binary functions. 
We plan to address this hypothesis in future works.
With Figure \ref{sf:1} we went deeper in our investigation and observed that also different implementation of the same functions actually lie closer in the vector space, as, for instance, the \emph{triangle} cluster, containing binaries for the {\sf gadd} function (an AES helper function) or the \emph{circle} and \emph{square} cluster, referring to the decryption and encryption function for the Camellia algorithm \cite{camellia}.  
Note that the gadd cluster contains not only different binaries obtained from the same source code, but also different implementations of the function. 
This behaviour (together with other observations not reported in the paper) seems to suggest that the network generalise further than our initial training assumption, where two functions are similar only if obtained from the same source code.


The second qualitative analysis was performed on the \textsf{i2v} embedding vectors for the ARM architecture. A similar analysis for the AMD64 architecture has been done in \cite{ChuaSSL17}.
From Figure \ref{fig:i2v_e} is possible to observe that the \textsf{i2v} model can actually incorporate both semantic and syntactic structure of the assembly language into the vector space.
In fact, similar instructions actually lie closer in the vector space. We also observed interesting geometrical properties in the vector space, often referred to as analogies (i.e., the famous \emph{king - man + woman = queen} in \cite{Mikolov}).
For instance, we observed\footnote{we report the vector that is nearest to the point given by the exact operation.} that:
\begin{enumerate}
\item \texttt{add r2,r2,1 \textbf{-} add r0,r0,1 \textbf{+} \\ sub r0,r0,1 \textbf{=} sub r2, r2, 1 }
\item \texttt{ldr r5, [sp, 8] \textbf{-} ldr r5,[sp,0xc] \textbf{+} str r5,[sp,0xc] \textbf{=} str r5, [sp, 8] }
\end{enumerate}  
This seems to suggest that the \textsf{i2v} model capture very well the semantic relationship between the increment and the decrement instructions as well as semantic structure of operations that store and load data from a register to memory or viceversa.

\begin{figure}[t!]
\centering
	\includegraphics[width=\linewidth]{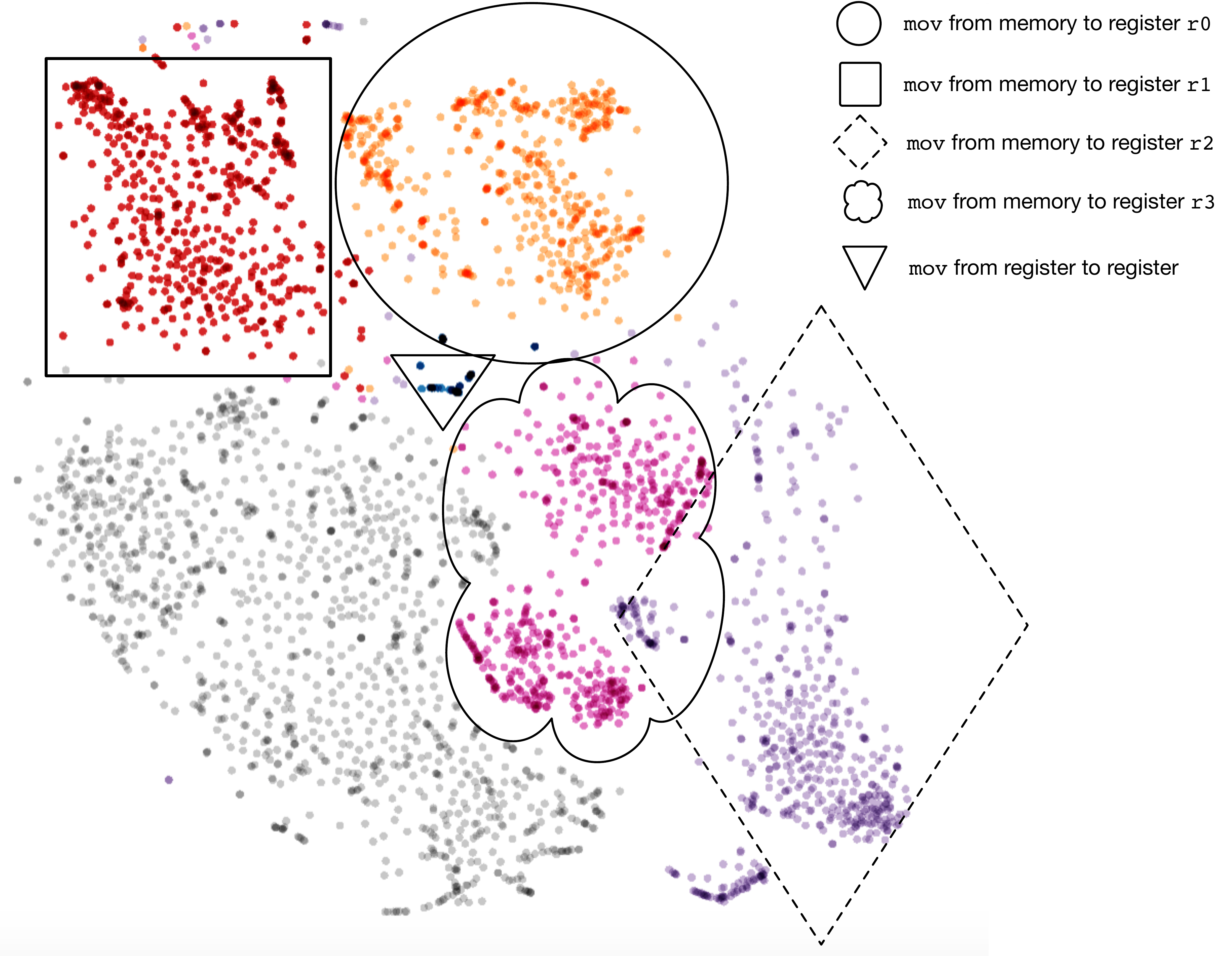}
\caption{2-dimensional visualisation of instruction embeddings for the mnemonics \texttt{mov}. The dimensionality reduction has been done using \emph{t-SNE}. 
The orange cluster, marked with a circle, is  the cluster of mov from memory to register \texttt{r0}.
The red cluster, marked with a rectangle,  is the cluster of mov from memory to register \texttt{r1}. 
The purple cluster, marked with a parallelogram, is  the cluster of mov from memory to register \texttt{r2}.
The pink cluster, marked with a cloud, is  the cluster of mov from memory to register \texttt{r3}.
The blu cluster, marked with a triangle, is the cluster of mov from register to register.
}
\label{fig:i2v_e}
\end{figure}

\section{Conclusions and future work}
\label{sec:conc}

In this paper, we proposed and investigated several solutions to create an annotated control flow graph 
by associating a feature vector for each vertex in the CFG.
This is the first step needed by the graph embedding neural network to generate binary function embedding.
In particular, we showed through an experimental campaign conducted on a state-of-the-art dataset that an unsupervised approach for feature learning leads to a boost in performance by almost $2\%$ with respect to current solutions based on manual feature engineering.
Our best performing solution (i.e., \wTv) make use of instruction embedding models, learned on a large corpus of binary instructions, and associate a feature vector to a vertex by computing a weighted average of all embedding vectors for the instructions contained in the vertex.
Our experimental results also showed that other feature extraction solutions, such as \sampTv, performs poorly and are probably not suited for the binary similarity task. 
We additionally performed some qualitative analysis on functions embedding that showed several insights worth mentioning. We believe that the clusters in Figure \ref{sf:2} are interesting and highlight possible lines for future work.
Finally, we have shown that the word2vec model can be adapted from the task of embedding natural language to the one of embedding assembly instructions. However, we believe that the assembly language has peculiarities that are not present in natural language.
For instance, the assembly code can be symbolically executed carrying a semantic that cannot be seen by a purely textual analysis. For this reason, we think that there is room for improvement in the instruction embedding model. For instance, the learning procedure could take into account both the nearby instructions, and information obtained by a symbolic execution of the instruction context.

\bibliographystyle{IEEEtranS}
 \bibliography{mybibfile}  

\end{document}